\documentclass[letterpaper, 10 pt, conference]{ieeeconf}  % Comment this line out if you need a4paper

\IEEEoverridecommandlockouts                              % This command is only needed if 
\usepackage{graphics} % for pdf, bitmapped graphics files
\usepackage{epsfig} % for postscript graphics files
\usepackage{mathptmx} % assumes new font selection scheme installed
\usepackage{times} % assumes new font selection scheme installed
\usepackage{amsmath} % assumes amsmath package installed
\usepackage{amssymb}  % assumes amsmath package installed
\usepackage[T1]{fontenc}
\usepackage{amsfonts}
\usepackage{booktabs}
\usepackage{siunitx}
\usepackage{caption} 
\usepackage{algorithm}
\usepackage{algorithmic}
\usepackage[flushleft]{threeparttable}
\title{\LARGE \bf
	Deep-Reinforcement-Learning-Based Semantic Navigation of Mobile Robots in Dynamic Environments
}

\author{Linh K\"astner$^{1}$, Cornelius Marx$^{1}$ and Jens Lambrecht$^{1}$% <-this % stops a space
	\thanks{$^{1}$Linh Kästner, Cornelius Marx and Jens Lambrecht are with the Chair Industry Grade Networks and Clouds Department, Faculty of Electrical Engineering, and Computer Science,
		Technical University of Berlin, Berlin, Germany
		{\tt\small linhdoan@tu-berlin.de}}%
}

\begin{document}

	\maketitle
	\thispagestyle{empty}
	\pagestyle{empty}

	%%%%%%%%%%%%%%%%%%%%%%%%%%%%%%%%%%%%%%%%%%%%%%%%%%%%%%%%%%%%%%%%%%%%%%%%%%%%%%%%
	\begin{abstract}
Mobile robots have gained increased importance within industrial tasks such as commissioning, delivery or operation in hazardous environments. The ability to autonomously navigate safely especially within dynamic environments, is paramount in industrial mobile robotics. Current navigation methods depend on preexisting static maps and are error-prone in dynamic environments. Furthermore, for safety reasons, they often rely on hand-crafted safety guidelines, which makes the system less flexible and slow. Visual based navigation and high level semantics bear the potential to enhance the
safety of path planing by creating links the agent can reason about for a more flexible navigation. On this account, we propose a reinforcement learning based local navigation system which learns navigation behavior based solely on visual observations to cope with highly dynamic environments. Therefore, we develop a simple yet efficient simulator - ARENA2D - which is able to generate highly randomized training environments and provide semantic information to train our agent. We demonstrate enhanced results in terms of safety and robustness over a traditional baseline approach based on the dynamic window approach.

	\end{abstract}

	%%%%%%%%%%%%%%%%%%%%%%%%%%%%%%%%%%%%%%%%%%%%%%%%%%%%%%%%%%%%%%%%%%%%%%%%%%%%%%%%
	\section{Introduction}
The demand for mobile robots has raised significantly due to their flexibility and the variety of use cases they can operate in. Tasks such as provision of components, transportation, commissioning or the work in hazardous environments are increasingly being executed by such robots \cite{smartfactory}, \cite{mobilerob}. A safe and reliable navigation is essential in operation of mobile robotics. Typically, the navigation stack consists of self-localization, mapping, global and local planning modules. Simultaneous Localization and Mapping (SLAM) is most commonly conducted as part of the navigation stack. It is used to create a map of the environment using its sensor observations upon which the further navigation relies on. However, this form of navigation depends on the preexisting map and its performance degrades at highly dynamic environments \cite{sun2017improving}, \cite{bahraini2019slam}. Furthermore, it requires an exploration step to generate a map which can be time consuming especially for large environments.\\
Deep Reinforcement Learning (DRL) emerged as a solid alternative to tackle this challenge of navigation within dynamic environments \cite{zeng2019navigation}. A trial and error approach lets the agent learn its behavior based purely on its observations. The training process is accelerated with neural networks and recent research showed remarkable results in mobile navigation. Yet, a main concern still lies in the safety aspect of navigation within human robot collaboration. Most ubiquitous are hand defined safety restrictions and measures which are non flexible and result in slow navigation. Higher level semantics bear the potential to enhance the safety of path planing by creating links the agent can reason about and consider for its navigation \cite{borkowski2010towards}. On this account, we propose a deep DRL local navigation system for autonomous navigation in unknown dynamic environments that works both in simulation and reality. To alleviate the problem of overfitting, we include highly random and dynamic components into our developed simulation engine called ARENA2D. For enhanced safety, we incorporate high level semantic information to learn safe navigation behavior for specific classes. The result is an end to end DRL local navigation system which learns to navigate and avoid dynamic obstacles based directly on visual observations. The robot is able to reason about safety distances and measures by itself based solely on its visual input. The main contributions of this work are following:
	\begin{itemize}
		\item Proposal of a reinforcement learning local navigation system for autonomous robot navigation based solely on visual input.
		\item Proposal of an efficient 2D simulation environment - ARENA2D - to enable safe and generalizable navigation behavior.
		\item Evaluation of the performance in terms of safety and robustness in highly dynamic environments.
	\end{itemize}
	The paper is structured as follows. Sec. II gives an overview of related work. Sec. III presents the conceptional design of our approach while sec. IV describes the implementation and training process. Sec. V demonstrates the results. Finally, Sec. VI gives a conclusion.
	
	\section{RELATED WORK}
	
	\subsection{Deep Reinforcement Learning for Navigation}
	
	%Reinforcement Learning for navigation tasks gained increased interest to provide a more flexible approach compared to traditional approaches which often rely on maps. Early works include Kohl et al.  \cite{kohl2004policy} who proposed a policy gradient RL algorithm to optimize the behavior of an autonomous robot for navigation tasks.  Vasquez et. al \cite{vasquez2014inverse} and Ziebart et. al \cite{ziebart2008maximum} used inverse reinforcement learning to generate models for navigation behavior. However, the training requires a tedious acquisition of suitable and efficient policies.\\
	With the advent of powerful neural networks, deep reinforcement learning (DRL) mitigated the bottleneck of tedious policy acquisitions by accelerating the policy exploration phase using neural networks. Mhni et. al \cite{mnih2015human} first used neural networks to find an optimal policy for navigation behavior. They conducted high level sensory input and proposed an end-to-end policy learning system termed Deep-Q-Learning (DQN). 
	Bojarski et. al \cite{bojarski2016end} applied the same techniques for mobile robot navigation by proposing an end-to-end method that maps raw camera pixels directly to steering commands within a simulation environment and show the feasibility of a RL approach. Most recently, Pokle et al. \cite{pokle2019deep} presented a system for autonomous robot navigation using deep neural networks to map observed Lidar sensor data to actions for local planning. 
	Zeng et. al \cite{zeng2019navigation} proposed a DRL system to work in with unknown dynamic environments. The researchers include a gated recurrent unit to interact with the temporal space of observations and train the agent with moving obstacles in simulation. They show the feasibility and efficiency of the method in terms of navigation robustness and safety in dynamic environments. Nevertheless, the method is still limited to simulation. In contrast to that, our work proposes a system trained in unknown dynamic environments and additionally, transfers the algorithms into the real robot. 
	
\subsection{Semantic Navigation}
Semantic navigation have been a research direction for many years. 
Borowski et al. \cite{borkowski2010towards} introduces a path-planning algorithm after semantically segmenting the nearby objects in the robot’s environment. The researchers were able to show the feasibility and extract information about object classes to consider for mobile robot navigation.
Wang et al. \cite{wang2018visual} propose a three layer perception framework for achieving semantic navigation. The proposed network uses visual information from a RGB camera to recognize the semantic region the agent is currently located at and generate a map.  
The work demonstrates that the robot can correct its position and orientation by recognizing current states from visual input. However, the dynamic elements such as presence of dynamic pedestrians are not taken into consideration.
Zhi et al. \cite{zhi2019learning} proposes a learning-based approach for semantic interpretation of visual scenes. The authors present an end-to-end vision-based exploration and mapping which builds semantics maps on which the navigation is based. One limitation of their method is the assumption of a perfect odometry, which is hard to achieve in real robots.
Zhu et al. \cite{zhu2017target} presented a semantic navigation framework where the agent was trained with 3D scenes containing real world objects. The researchers were able to transfer the algorithm towards the real robot which could navigate to specific real world objects like doors or tables. Furthermore, they showed the potential of using semantics for navigation without any hand crafted feature mapping but working solely on visual input. The training within the 3D environment, however, is resource-intensive and require a large amount of computational capacity. 
Our work follows a more simplified way of incorporating semantic information. We include different object classes into our 2D simulation environment and train the agent with specific policies which should shape the behavior of the robot when encountering with the specific classes. Furthermore, we transfer the algorithms towards the real robot.

\section{CONCEPTUAL DESIGN}
We propose a DRL-based local navigation system which maps observed sensory input directly to robot actions for obstacle avoidance and safe navigation in dynamic environments. Furthermore, we aspire to explore the safety enhancements using semantic information. More specifically, rules based on detected nearby objects are defined and incorporated into the reward functions: the mobile robot have to keep a distance of 0.8 meters from detected humans and 0.3 meters from collaborating robots. The training is accelerated with neural networks and build upon a \textit{deep Q network} (DQN) which maps states to a so called Q values that each state action pair possess. Therefore, we employ the proposed DRL workflow described in \cite{10.5555/3279266}.

\subsection{Design of the Proposed Navigation System}
\begin{figure}[]
		\centering
		\includegraphics[width=3.3in]{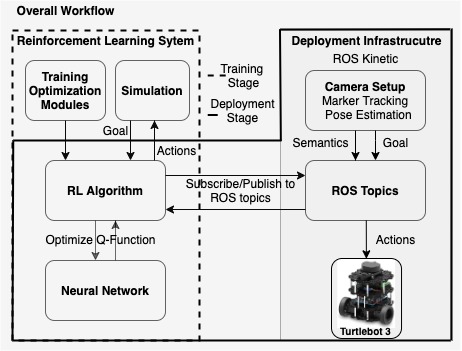}
		\caption{Design of the navigation system}
		\label{concept}
	\end{figure}

The general workflow of our system is illustrated in Fig. \ref{concept}. The agent is trained by the RL algorithm within simulation by analyzing states, actions and rewards.
For our use case, the states are represented by laser scan observations and object detections while actions are the possible robot movements. Within the training stage, these information are simulated in our simulation environment. Within deployment stage, a RGB camera with integrated object detection and a 360 degree 2D Lidar scanner of the Turtlebot3 deliver the required data.
Core component of our system is the neural network which optimizes the Q function.
We input the data into the neural network which maps the states-actions tuples to Q values with the maximal reward. Thus, the agent learns optimal behavior given a set of states and actions. In addition, we integrate several optimization modules to accelerate the training process even further. Finally, in the deployment stage, the RL model is deployed towards the real robot using a proposed deployment infrastructure consisting of a top down camera to detect the goal and surrounding objects. As a middle-ware for communication between all entities, ROS Kinetic on Ubuntu 16.04 is used. 
 For the deployment of the trained algorithms to the real robot, several challenges have to be considered. Unlike simulation, the robot does not know when the goal is reached. Therefore, we propose a solution using object detection and markers with the aforementioned global camera.
The problem of inaccurate and noisy laser scan data is considered within our simulation environment, where we include a module to add several levels of noise to the laser scan data. Furthermore, we included dynamic obstacles and randomness into the simulation to alleviate the differences between real and simulation environment.

\section{IMPLEMENTATION}
In the following chapter, each module of our proposed navigation system is presented in detail.
	
\subsection{Training Algorithm }
The basic training loop is based on the suggestions from \cite{10.5555/3279266} and employs deep Q-learning. 
We rely on three major techniques: \textit{replay buffer} and \textit{epsilon greedy} to cope with the exploration and exploitation problem and \textit{target net} which we use to stabilize the training process in terms of robustness. To abstract the algorithm for our simulation environment, we split it into two parts: a simulation step, which interacts with the simulation environment, and a learning step, which is responsible for the refinement of the neural network model. The implementation of those steps is described by algorithms \ref{alg:PreStep} and \ref{alg:PostStep}. 
\begin{algorithm}[]
		\caption{PreStep}
		\label{alg:PreStep}
		\begin{algorithmic}
			%\STATE $Q \leftarrow \mbox{random\_weights()}$
			%\STATE $Q_T \leftarrow \mbox{random\_weights()}$
			%\STATE $s \leftarrow \mbox{simulation\_reset}()$ 
			%\STATE $R \leftarrow$ new FIFO Queue$(R_{\mbox{size}})$
			%\STATE $\bar{r} \leftarrow 0$
			\IF{$\mbox{random}() < \epsilon$}
			\STATE $a \leftarrow \mbox{random\_action}()$
			\ELSE
			\STATE $a \leftarrow arg\max\limits_{a}Q(s,a)$
			\ENDIF
			\STATE $(s', r) \leftarrow \mbox{simulation\_step}(a)$
			\IF{episode is over}
			\STATE $R.\mbox{insert}((s, s', a, r, \mbox{TRUE}))$
			\STATE $s \leftarrow \mbox{simulation\_reset}()$
			\ELSE
			\STATE $R.\mbox{insert}((s, s', a, r, \mbox{FALSE}))$
			\STATE $s \leftarrow s'$
			\ENDIF
			\STATE $t \leftarrow t + 1$
		\end{algorithmic}
	\end{algorithm} In the \textit{PreStep} algorithm, a random action is chosen with a probability of $\epsilon$. Otherwise, the action with the maximum Q Value, according to the current networks estimation, is retrieved. Using that action, a simulation step is performed, revealing the new state and the reward gained. The new state along with the previous state, reward, action and a flag indicating, whether the episode has ended or not, is stored in the replay buffer. \\
	The actual training takes place in the \textit{PostStep} algorithm. Here a random batch is sampled from the replay buffer and the mean square error (MSE) loss is calculated using the Bellman equation.
	Every $N_{\mbox{sync}}$ frames, weights from the network $Q$ are copied over to the target network $\hat{Q}$.
	Using stochastic gradient descent (SGD) optimization, the weights of the network $Q$ are optimized according to the MSE loss $L$ calculated for every batch sample. Finally, the epsilon value is updated according to the current step $t$. Thereby, epsilon denotes the randomness of executed actions in each step which ensures an efficient trade off between exploration and exploitation. 
	\textit{PreStep} and \textit{PostStep} are called in a loop until the network converges.
	
	\begin{algorithm}[]
		\caption{PostStep}
		\label{alg:PostStep}
		\begin{algorithmic}
			\STATE $B \leftarrow R.\mbox{random\_batch}(B_{\mbox{size}})$
			\STATE $L \leftarrow \mbox{new Array}(B_{\mbox{size}})$
			\FOR{$i$ in $B$}
			\STATE $(s_i, s'_i, a_i, r_i, d_i) \leftarrow B[i]$
			\IF{$d_i = \mbox{TRUE}$}
			\STATE $y \leftarrow r_i$
			\ELSE
			\STATE $y \leftarrow r_i + \gamma \max\limits_{a}\hat{Q}(s'_i, a)$
			\ENDIF
			\STATE $L[i] \leftarrow (Q(s_i, a_i) - y)^2$
			\ENDFOR
			\STATE $Q.\mbox{optimize}(L)$
			\IF{$t \bmod N_{\mbox{sync}} = 0$}
			\STATE $Q_T \leftarrow Q$
			\ENDIF
			\STATE $\epsilon \leftarrow \max(\epsilon_{min}, 1-t / t_{max})$
		\end{algorithmic}
	\end{algorithm}

%	\subsection{Optimization Modules}
%	By introducing two additional techniques, some improvements have been made regarding training stability and convergence speed. In addition, enhancements were made to fully utilize GPU hardware.
	
%	\subsubsection{\textbf{N-Step-DQN}}
%	Sutton et. al \cite{sutton1988learning} proposed the N-Step optimization to unroll the Bellmann equation for $N$ steps. For $N = 2$ this results in:
	
%	\begin{equation}
%	Q(s_t, a_t) =  r_t + \gamma r_{t+1} + \gamma^2\max\limits_{a}Q(s_{t+2}, a)
%	\end{equation}
	
%	This mainly improves the convergence speed, as it eases the possibility to maximize future rewards. However, it assumes that the optimal action was taken in every step, which is why too high values for $N$ will eliminate convergence all together. Therefore, we considered N $\in$ [1,2].
	
%	\subsubsection{\textbf{Double-DQN}}
%	Works from Van Hasselt et al. \cite{van2016deep} have shown that the DQN tends to overestimate the Q values resulting in suboptimal policies. Training dynamics can be improved when using the action with the maximum Q Value currently predicted by the network. Therefore, the discounted Q Value of the next step using the target network is calculated, instead of the actual action that was taken \cite{10.5555/3279266}:
%	\begin{equation}
%	y = r + \gamma\hat{Q}(s_{t+1}, arg\max\limits_{a}Q(s_{t+1},a))
%	\end{equation}

\subsection{Neural Network Design}
\begin{figure}[]
		\centering
		\includegraphics[width=3.3in]{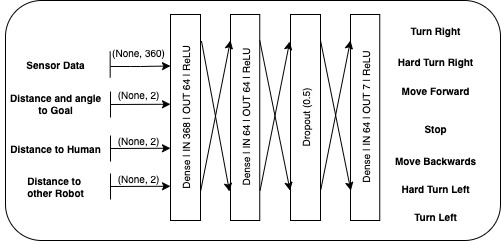}
		\caption{Architecture of fully connected neural network}
		\label{fc1}
\end{figure}

We use fully connected neural networks for our DQL sytem. Our model consists of 4 hidden and fully connected layers and is described in Fig. \ref{fc1}
	
The input of the first fully connected layer represents the laser scan data and information about nearby humans or robots. 360 neurons as input values representing one value for each degree of the laser scanner.
The nearby dynamic objects are each represented with two neuron using their distance and angle to the robot. For simplicity reasons, we restrict this input to one object for each class human and robot. 
After 2 dense layers and a dropout layer, the resulting output are 7 neurons denoting the robots actions. Adam is chosen as optimizer with an adaptive learning rate of 0.0025. As loss-function, Mean Squared Error (MSE) is chosen.
	
\subsection{Training Stage with ARENA2D }
\begin{figure}[H]
	\centering
	\includegraphics[width=3.3in, height=3in]{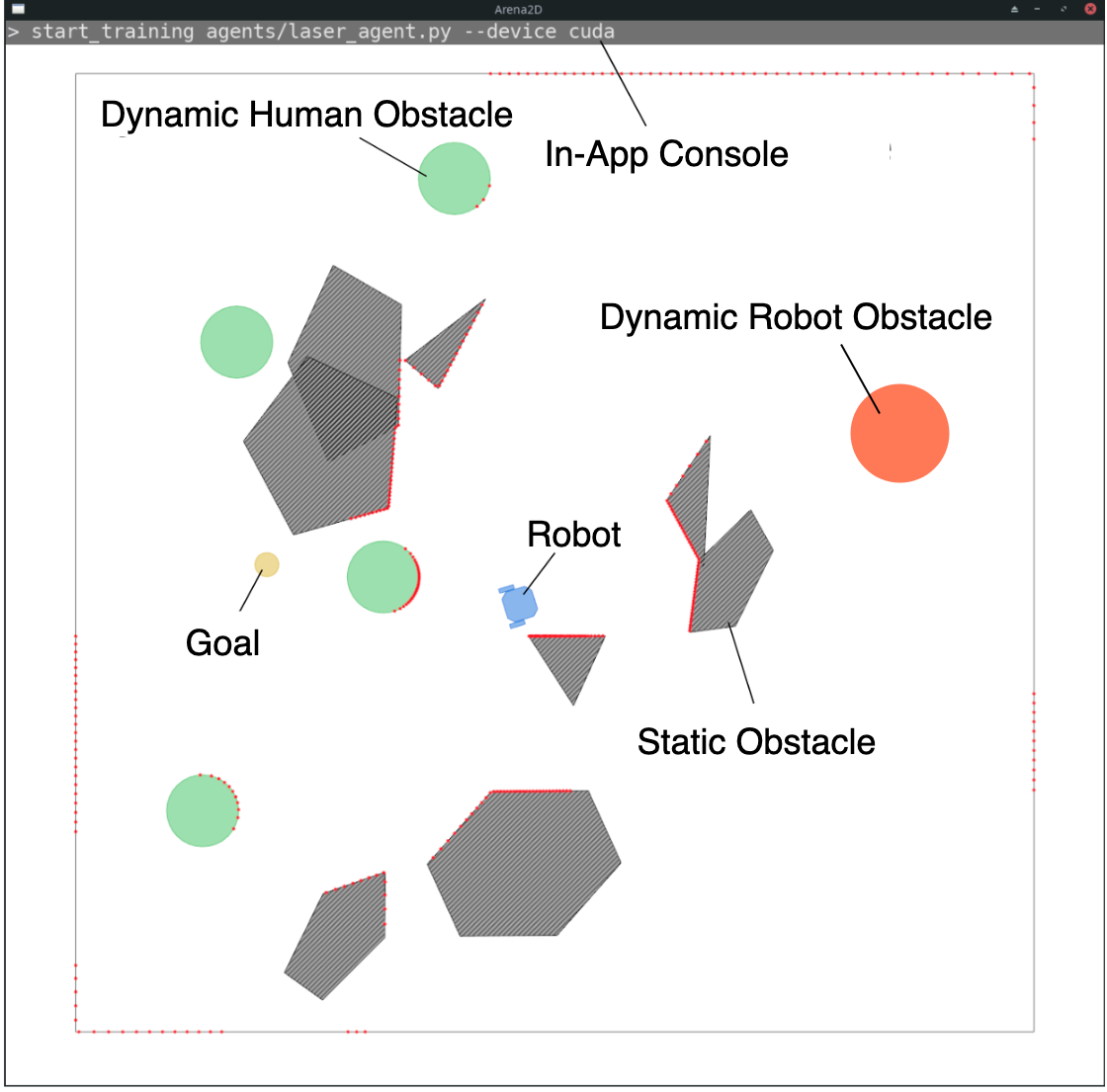}
	\caption{Graphical user interface of the simulation environment ARENA2D}
	\label{GUI}
\end{figure}

We apply the presented methods in the training stage of the robot through simulations to generate a model which, subsequently, can be used for the real robot. Therefore, we developed a simple yet efficient 2D training environment - ARENA2D - with a large amount of built-in capabilities for performance enhancements and exploration of training settings. The simulation environment is depicted in Fig. \ref{GUI}. Within the simulation environment, we include static as well as dynamic components and create several stages with ascending difficulty. In total, we executed the training on 3 different stages which we denote as static, dynamic and semantic. The static stage contains only static obstacles whereas the dynamic
stage include moving obstacles. These two stages use a neural network which does not include the neurons indicating the distance of human and collaborating robot as additional
input. The semantic stage uses the network presented in Fig. \ref{fc1}. If the robot hits a wall or times out, it is reseted to the center of the stage. For human obstacles, we included an additional stop rate which lets the obstacle stop at random positions for a time of 2 seconds thus simulating the human behavior. 
\subsubsection*{\textbf{Agent Definition}}
To simulate the real robot, we define an agent for simulation.
The agent is defined with the same parameters and output compared to the real robot in order to have as little differences as possible between simulation and reality. Therefore, we choose the \textit{Turtlebot3} due to its simplicity and compactness. It is equipped with a 360 degree laser scan sensor and offers input for further equipment e.g. an on board camera. The possible actions are listed in Table \ref{actions}.
\begin{table}[H]
		\centering
		\caption{Agent definition}
		\begin{tabular}{lllll}
			\hline
			Action  & Linear Velocity  [m/s]  & Angulat Velocity  [rad/s]    \\ \hline
			Forward &0.15        & 0          \\ 
			Backwards    & -0.15       &0         \\ 
			Stop    & 0       &0         \\ 
			Left    & 0.15        & + 0.75          \\ 
			Right    & 0.15       &- 0.75         \\ 
			Strong Left    & 0.15       &+ +1.5         \\ 
			Strong Right    & 0.15       &- 1.5         \\ 
			\hline
		\end{tabular}
		
		\label{actions}
	\end{table}
%\subsubsection*{\textbf{Training setup/ parameters}}
%We use several training parameter setups. Important parameters for each training run are listed in Table \ref{tabletrain}. 
%All models were trained on a GPU server with 2 NVIDIA GTX Ti 2046 Graphic cards of 11 GB and 64GB RAM until relevant metrics such as average reward or loss converged. We executed the training in the different static and dynamic training environments presented previously. 
%	\begin{table*}[]
%	\centering
%	\renewcommand{\arraystretch}{1.5}
%	\vskip0.2cm
%	\caption{Training Setups for agent training stage in simulation environment}
	
	%\begin{tabular}{c|cccccp{6.cm}}
	%	\hline
	%	Training Name    & Time [h] & SuccessRate [\%] &Episodes  & Environment & NN Structure&Techniques  \\ \hline
	%	DRL stat. &  5.5       & 99.78     & 130,000 & static &FC\footnote[1] & Epsilon Greedy, Target Network, Replay Buffer, 2-Step Evaluation, 2-Step-DQN Double-DQN \\  \cline{2-7}
	%	DRL dyn. &  27.9      & 99.95      & 200,000 & dynamic & FC\footnote[1]& Epsilon Greedy, Target Network, Replay Buffer, 2-Step Evaluation, 2-Step-DQN, Double-DQN \\  \cline{2-7}
	%	DRL sem. &  47.5      & 99.98      & 500,000 & dynamic & FC\footnote[1]& Epsilon Greedy, Target Network, Replay Buffer, 2-Step Evaluation, 2-Step-DQN, Double-DQN \\
	%	\hline
		
%	\end{tabular}
%	\begin{tablenotes}
	%	\item[1] $^1$ Fully Connected Neural Network.
%	\end{tablenotes}
	
%	\label{tabletrain}
	
%\end{table*}
	\subsubsection*{\textbf{Rewards and Penalties}}
The rewards were exactly the same for all trainings. After each step, the agent will receive a reward based on the new state the robot is in. $\alpha$ denotes the angle between the robot and the goal. The rewards and penalties are listed in Table \ref{tablerew}.
\begin{table}[htbp]
	\centering
	\caption{Rewards and penalties for training}
	\begin{tabular}{lllll}
		\hline
		Event & Description & Reward & Ep. Over\\
		\hline
		Goal reached & yes & $+100$ & yes\\
		Moving towards goal & $|\alpha| \leq 30\si{\degree}$ & $+0.1$  & no\\
		Wall hit & Robot hit wall & $-100$ & yes\\
		Moving away from goal & $|\alpha| > 30\si{\degree}$ & $-0.2$  & no\\
		Violate distance to human & $d < 0.7 m$ & $-10$  & no\\
		Violate distance to robot & $d <  0.2 m$ & $-10$  & no\\
		\hline
		
	\end{tabular}
	
	\label{tablerew}
	
\end{table}

	\subsubsection*{\textbf{Hyperparameters}}
To determine the optimal hyperparameters, we conducted several training runs and adjusted the hyperparameters manually according to our literature research as well as experience. The optimal hyperparameters used for all further training runs are listed in Table \ref{tablehyper}. 
 \begin{table}[htbp]
		
		\renewcommand{\arraystretch}{1.3}
		\caption{Hyperparameters for training}
		\begin{tabular}{ccp{4.1cm}}
			\hline
			Hyperparameter  &Value& Explanation    \\ \hline
			Mean Success Bound & 1        & Training considered done if mean success rate reaches this value           \\ 
			Num Actions   & 7        &Total number of discrete actions the robot can perform           \\ 
			Discount Factor & 0.99    & Discount factor for reward estimation (often denoted by gamma)   \\ 
			Sync Target Steps   & 2000      &Target network is synchronized with current network every X steps        \\ 
			Learning Rate   & 0.00025        &Learning rate for optimizer           \\ 
			Epsilon Start  & 1        &Start value of epsilon        \\ 
			Epsilon Max Steps   & $10^5$        &Steps until epsilon reaches minimum         \\ 
			Epsilon End   & 0.05        &Minimum epsilon value    \\ 
			Batch Size   & 64        &Batch size for training after every step          \\ 
			Training Start  & 64        &Start training only after the first X steps        \\ 
			Memory Size  & $10^6$        &Last X states will be stored in a buffer (memory), from which the batches are sampled        \\ 
			\hline
		\end{tabular}
		
		\label{tablehyper}
		
	\end{table}

\subsection{Deployment on real robot}
\begin{figure}[]
	\centering
	\includegraphics[width=3.3in, height=2.9in]{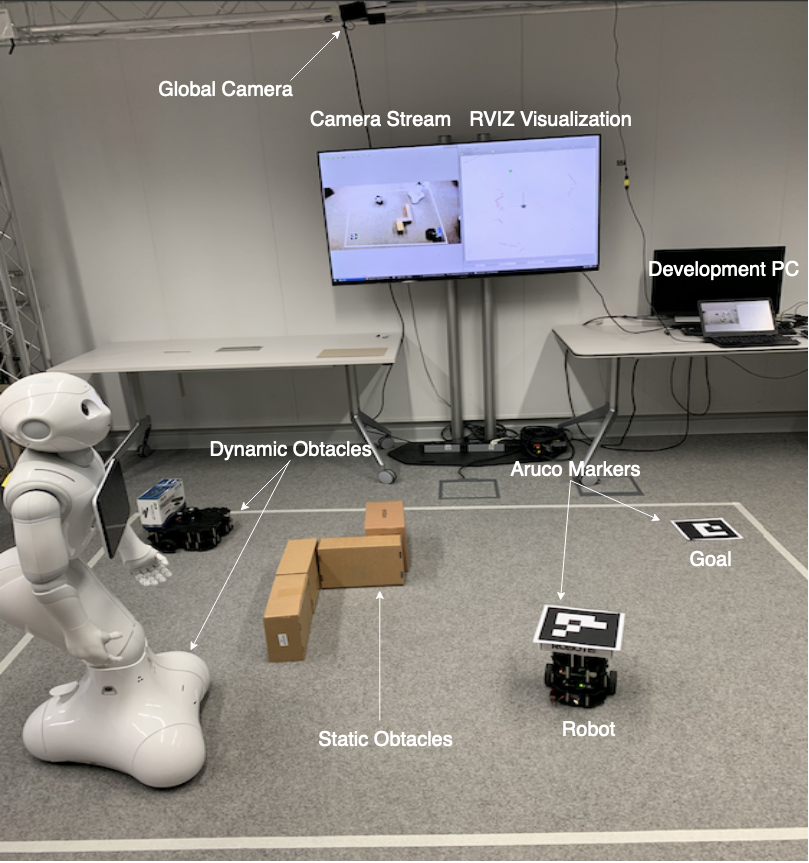}
	\caption{Deployment setup}
	\label{re}
\end{figure}

Once the simulation was successful and the agent performs a safe navigation within all simulation environments, we deploy the algorithms towards the real robot to evaluate their feasibility within the real environment. Fig. \ref{re} illustrates the deployment setup with all entities used for conducting the experiments. Fiducial Aruco markers \cite{romero2018speeded} are included on the robot and the goal to verify the arrival at the target destination by comparing the position of both markers. Therefore, all entities are tracked with a global Intel Realsense D435 camera placed at the top of the real test environment. When both markers reach the same position, the system is informed that the destination is reached. The robot is equipped with an Intel Realsense camera as well, which delivers input for the human pose estimation module. The communication and signal workflow between all entities is explained in the following chapter. A variety of different obstacles were included, which are similar to the simulation as well as completely different. For static obstacles, chairs, round objects and boxes of different sizes and forms were used. Dynamic obstacles include other robots moving randomly and moving humans walking randomly across the environment and intersecting the path of the robot.

%\subsection*{Workflow and Signal Pipeline}

%\begin{figure}[!h]
%%%	\label{pipeline2}
%\end{figure}
%Communication between all entities is established and controlled via asynchronous ROS communication nodes. Fig. \ref{pipeline2} illustrates the relevant topics and their signal workflow. 
%The system consists of 5 ROS nodes: the \textit{Turtlebot3} node running on the robot and the \textit{Agent-, Marker Detection-, Pose Estimation- and Observation Packer} nodes running on the GPU server. The \textit{Observation Packer} node is regarded as the core gateway entity to orchestrate all communication between the entities. The \textit{Turtlebot3} node is publishing the data of the laser scanner and odometry sensors directly to the \textit{Observation Packer}. 
%The \textit{Pose Estimation} node subscribes to the camera stream of the onboard camera and publishes the detections to the \textit{Observation Packer}.
%Equivalently, the  \textit{Marker\_detector} node subscribes to the top down global camera stream and observes the relative position and orientation between the robot and the goal, by detecting \textit{Aruco} markers. Subsequently, it publishes the relative position to the \textit{Observation Packer}. It receives all incoming observations, from the aforementioned nodes, preprocesses them and publishes the data to the observation topic. 
%The $agent$ node receives \textit{observation} messages, which are forwarded through the trained neural network to retrieve an action. That action is translated into a linear and angular velocity and published to the topic \textit{cmd\_vel} which activates the motor drivers of the \textit{Turtlebot3}. To ensure robustness due to an unstable marker detection or when the robot is outside the top down cameras view, we include odometry data inside the \textit{observation\_packer} node to update the relative position of the goal accordingly.

\subsection{Pose Estimation}
For the object detection module, we utilized a pose estimation module working with RGB input based on SSPE \cite{tekin2018real}. Thus, the position and distance of humans or collaborating robots can be detected globally. Currently, the model is able to localize humans, the \textit{Kuka Youbot} and the \textit{Turtlebot} models \textit{Burger} and \textit{Waffle}. We fine-tune the model with a training on a human and robot RGB-dataset utilizing the pipeline proposed in our previous work \cite{kastner2020markerless}. The results are transmitted to the \textit{Observation Packer} node to be considered for the agent. Subsequently, the DRL algorithm will refine the trajectory of the robot. 

%The process is illustrated in Fig. \ref{semres}.

%		\begin{figure}[]
%		\centering
%		\includegraphics[width=3.3in, height=2.9in]{semres}
%		\caption{Object detection of the top down camera}
%		\label{semres}
%	\end{figure}

The representing neuron for the distance of the agent to humans and other robots is initially set to a distance of 10 meters to make sure the neural network is assuming no nearby human or robot. Once a robot or human is detected and localized, the estimated position will be given as input to the neural network via the \textit{Pose Estimation} node.

\subsection*{Conducted Experiments}
\begin{figure}[]
		\centering
		\includegraphics[width=3.3in]{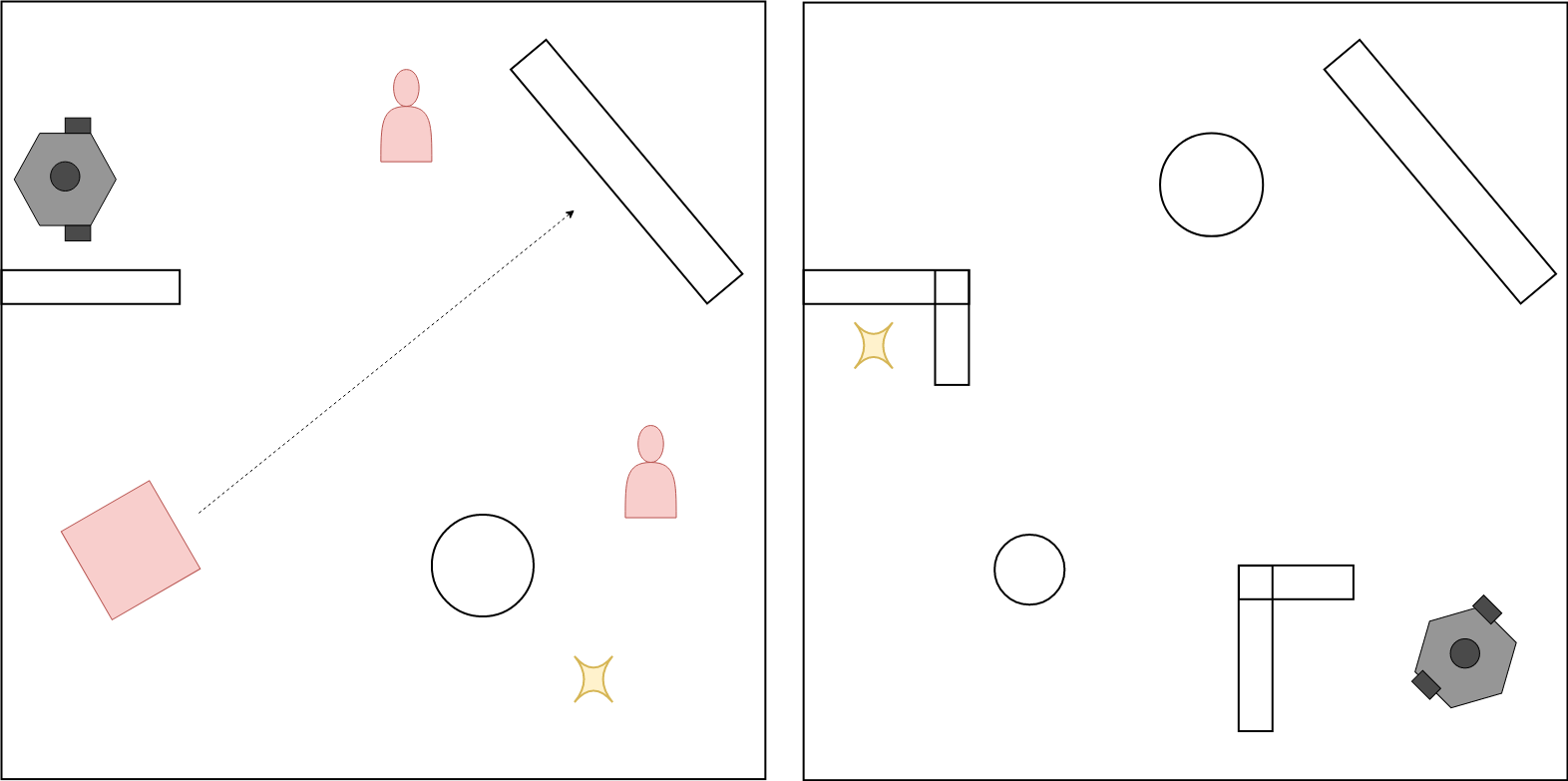}
		\caption{Test scenarios for the conducted experiments}
		\label{rese}
	\end{figure}

We conduct several experiments with the model at different real environments to compare our method against the traditional local planer of the \textit{Turtlebot3} navigation package that uses a preexisting map of the environment with static obstacles and an algorithm without semantic rules. The setup of the experiments are illustrated in Fig. \ref{rese}
We tested different setups with static as well as dynamic components like moving humans and other robots and placed 10 different goal positions ranging from 0.2m to 2.5m distance. The start position of the robot was the same for every run. For each approach, we conducted 30 measurements consisting of 3 measurements for each of the 10 goals. If the robot could not reach its target within a time of 1 minute or due to a shut down of the \textit{Turtlebot3} navigation planners because no path could be calculated, we increased the failure count but conducted another measurement to ensure that each approach has the same number of measurements. The mentioned shut down happens, if the navigation package can not localize the robot and fail to generate a path due to a too distant or complex goal or sudden obstacles interfering which at times result in a shut down.
	
	To explore the efficiency of the additional semantic rules, we compared the collision count of each of the approaches to reason about the safety of each approach. 
	
	\section{Results and Evaluation} 
	In the following, the results form our conducted experiments are presented.
	The deployment of the models to the real robot was without any difficulties and we compare our approaches with the traditional local planer of the \textit{Turtlebot3} navigation package in terms of relevant metrics such as distance speed, error rate and safety of the navigation. The results are listed in Table \ref{tableres}. The distance metric is indicating the efficiency of the path planer and is conducted through the odometry topic. We measured the time each approach required to reach the goal. The safety rate is calculated with the total number of collision the robot had with static or dynamic obstacles while still reaching the goal. Robustness describes how many times the robot failed to reach the goal due to a failed path planning resulting in a navigation stack shut down or if the robot pursuits a completely wrong direction and were out of the arena. In total we placed 10 different goal positions and for each goal, 3 measurements were carried out for all approaches. If one run resulted in a failure, this was added to the error rate count and another measurement was conducted such that finally, there are 30 measurements for each approach to calculate the mean distances and speed. The error rate is calculated as the percentage of failed to successful runs. Table \ref{tableres} lists the the results for each approach.
	\begin{table}[!h]
		\centering
		\caption{Comparison of navigation approaches }
		\begin{tabular}{lllll}
			\hline
			Metric   & Trad. &DRL Stat. & DRL Dyn. & DRL Sem.  \\ \hline
			Distance [m] & 4.72  &  3.71    & \textbf{4.1}         & 5.28         \\ 
			Speed  [s]  & 15.7      &\textbf{11.7}  & 12.49      & 17.9       \\ 
			Error Rate [\%]    & 16.66 &10       &  3.33        & \textbf{0}       \\ 
			Obstacles hit  &6.4   &  4.8 & 3.2   & \textbf{0} \\ \hline
		\end{tabular}
		
		\label{tableres}
		
	\end{table}
	It can be observed that our approaches outperform the traditional local planer of the robot in terms of speed and distance. Furthermore, our methods eliminate the need to generate a map which is necessary for the SLAM packages on which the global and local planner of the robot rely on.
	The model trained with semantic information achieves the best performance in terms of obstacle avoidance and had no collisions in all our test runs. Although this comes at the cost of longer distances and speed because the robot will keep a larger distance when encountering a human sometimes driving backwards. The greater distance alleviates the high amount of collisions that were observed in our previous work where we mitigated the issue by training in highly dynamic environments. The additional semantic information enhances this effect even further as indicated in table \ref{tableres}. 
	Notably, the transfer of the simulated agent to the real environment did not cause major differences, even though in the simulation environment, only round obstacles were deployed as dynamic obstacles. However, our agent could generalize its behavior to all obstacles both static and dynamic, thus still managed to avoid the objects and keep a save distance to the human. For a more visual demonstration of our experiments we refer to our demonstration video which is available at https://youtu.be/KqHkqMqyStM.

	\section{CONCLUSION}
	We proposed an overall deep reinforcement learning based local navigation system for autonomous robot navigation within highly dynamic environments. Therefore, we developed a simple, yet efficient simulation engine from scratch which showed fast and efficient training speed and feasibility in transferring the models towards the real robot. Our navigation algorithm works solely on visual input and eliminates the need for any additional map. Furthermore, we explored the potential of semantic information by incorporating semantic classes such as human and robot and concluded safety enhancements for the navigation. This will be extended in our further work to include more classes such as long corridor, doors or restricted areas. For the deployment into the real environment a framework was proposed to integrate the DRL algorithms towards the real robot using marker detection and odometry data. Thereby, we ease the transferability of simulated models and enable a map-independent solution.
	The results were remarkable both in static as well as dynamic environments and surpasses the traditional baseline RRT-PWA  approach in terms of safety and robustness. For future work, we plan to incorporate more semantic classes such as long corridor, doors or restricted areas into the training environment to enhance safety and the overall performance even further. Additionally, an extension of our framework for more capabilities and features e.g. including more reinforcement learning algorithms, recurrent modules and continuous actions is planned.

	\addtolength{\textheight}{-2cm}   % This command serves to balance the column lengths
	
	%%%%%%%%%%%%%%%%%%%%%%%%%%%%%%%%%%%%%%%%%%%%%%%%%%%%%%%%%%%%%%%%%%%%%%%%%%%%%%%%

	%%%%%%%%%%%%%%%%%%%%%%%%%%%%%%%%%%%%%%%%%%%%%%%%%%%%%%%%%%%%%%%%%%%%%%%%%%%%%%%%

	%%%%%%%%%%%%%%%%%%%%%%%%%%%%%%%%%%%%%%%%%%%%%%%%%%%%%%%%%%%%%%%%%%%%%%%%%%%%%%%%

	%%%%%%%%%%%%%%%%%%%%%%%%%%%%%%%%%%%%%%%%%%%%%%%%%%%%%%%%%%%%%%%%%%%%%%%%%%%%%%%%
	
	\bibliographystyle{IEEEtran}
	
	\bibliography{references}

\end{document}